\documentclass[11pt,a4paper]{article}
\usepackage[hyperref]{emnlp2020}
\usepackage{graphicx}
\usepackage{tabularx}
\usepackage{times}
\usepackage{latexsym}

\usepackage[linesnumbered,ruled,vlined]{algorithm2e}
\usepackage{booktabs}
\usepackage{fancyvrb}
\usepackage{graphicx}

\usepackage{microtype}

\usepackage{cleveref}

\aclfinalcopy 
\def\aclpaperid{1947} 

\usepackage[colorinlistoftodos,prependcaption,textsize=tiny]{todonotes}
\usepackage{xspace}
\usepackage{xargs} 

\newcommandx{\ignore}[2][1=]{}
\newcommandx{\TODO}{{\color{red} TODO}\xspace}

\newcommand{\dialogue}{\textsc{Dialogue}\xspace}

\newcommand{\feedback}{\textsc{Feedback}\xspace}
\newcommand{\personachat}{\textsc{PersonaChat}\xspace}

\newcommand{\biencoder}{\textsc{BiEncoder}\xspace}
\newcommand{\polyencoder}{\textsc{PolyEncoder}\xspace}
\newcommand{\nofeed}{\textsc{NoFeedback}\xspace}
\newcommand{\feeback}{\textsc{Feedback}\xspace}
\newcommand{\heuristic}{\textsc{Heuristic}\xspace}
\newcommand{\feedresp}{\textsc{Feed2Resp}\xspace}
\newcommand{\remove}{\textsc{Remove}\xspace}
\newcommand{\retain}{\textsc{Retain}\xspace}
\newcommand{\rewrite}{\textsc{Rewrite}\xspace}

\SetKwInput{KwInput}{Input}                
\SetKwInput{KwOutput}{Output} 
\SetKwRepeat{Repeat}{repeat}{until}

\title{Learning Improvised Chatbots from 
\\ Adversarial Modifications of Natural Language Feedback}

\author{Makesh Narsimhan Sreedhar \\
  Mila \\
  Universite de Montreal \\
  \texttt{narsimma@mila.quebec} \\\And
 Kun Ni \\
 Mila \\
  Universite de Montreal \\
  \texttt{nikun@mila.quebec} \\\And
Siva Reddy \\
 Facebook CIFAR AI Chair, Mila \\
  McGill University \\
  \texttt{siva@cs.mcgill.ca} \\}

\date{}

\begin{document}
\maketitle
\begin{abstract}

The ubiquitous nature of chatbots and their interaction with users generate an enormous amount of data. Can we improve chatbots using this data? A self-feeding chatbot improves itself by asking natural language feedback when a user is dissatisfied with its response and uses this feedback as an additional training sample. However, user feedback in most cases contains extraneous sequences hindering their usefulness as a training sample. In this work, we propose a generative adversarial model that converts noisy feedback into a plausible natural response in a conversation. The generator's goal is to convert the feedback into a response that answers the user's previous utterance and to fool the discriminator which distinguishes feedback from  natural responses. We show that augmenting original training data with these modified feedback responses improves the original chatbot performance from 69.94\% to 75.96\% in ranking correct responses on the \personachat dataset, a large improvement given that the original model is already trained on 131k samples.\footnote{Our code is released at \url{https://github.com/ekunnii/adversarial-feedback-chatbot/}}  

\end{abstract}

\section{Introduction}
Enabling chatbots to indulge in engaging conversations requires massive datasets of human-human conversations \cite{ritter2011data,sordoni2015neural,vinyals2015neural,zhang2018personalizing,zhang2019dialogpt}.
Training such dialog agents requires substantial time and effort expended in the collection of adequate number of high quality conversation samples.

\citet{hancock2019learning} alleviate this problem by introducing a self-feeding chatbot which can directly learn from user interactions.
This chatbot requests users to provide natural language feedback when the users are dissatisfied with its response.

\citet{hancock2019learning} treat this feedback as a gold response to the wrong turn and use it as an additional training sample to improve the chatbot.

\begin{figure}
    \centering
    \includegraphics[width=\columnwidth]{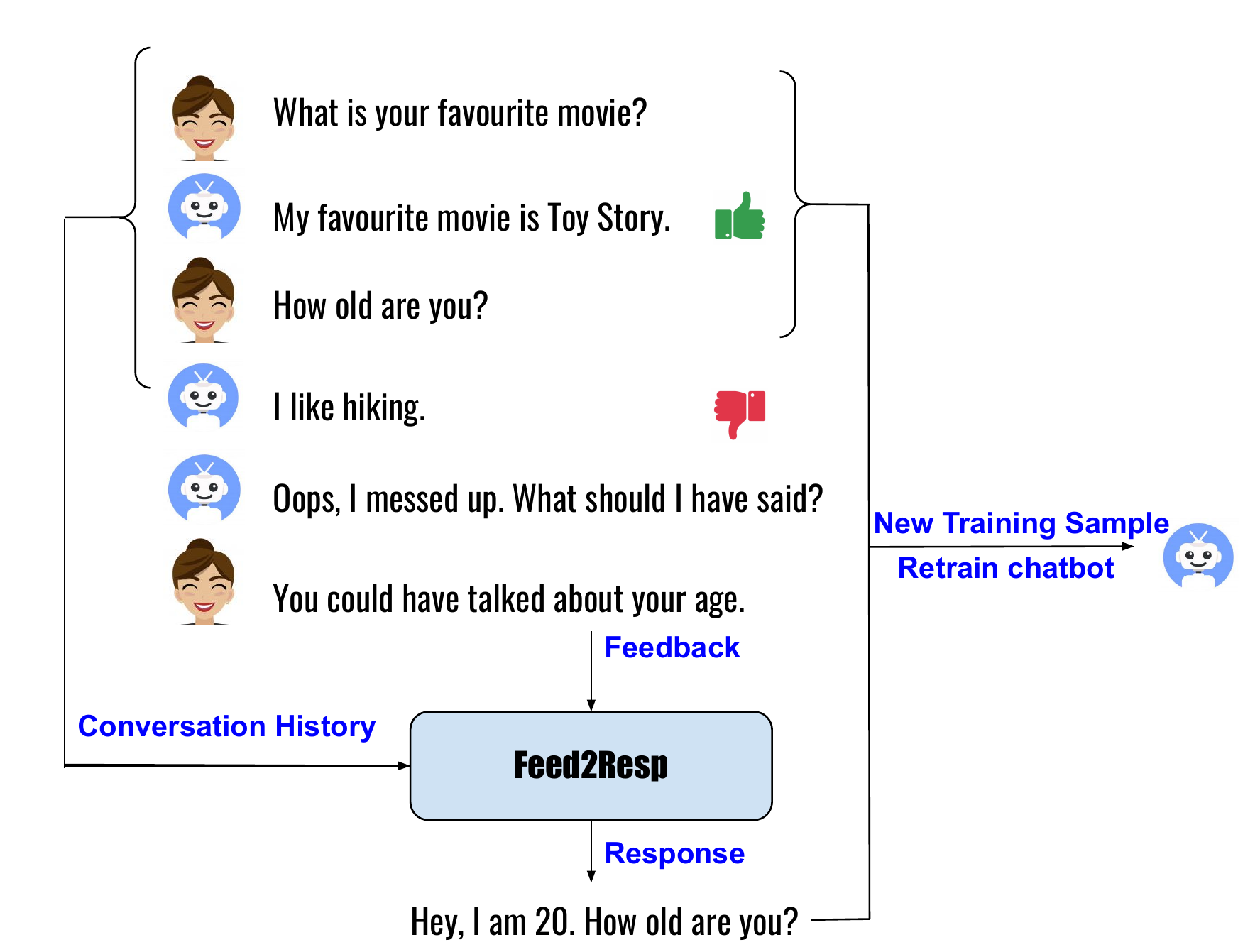}
    \caption{
        When the bot provides a poor response to the question posed by the user, the bot requests natural language feedback. 
        We use the conversation context and the feedback to construct a plausible response to the user query and use it as an additional training sample to improve the chatbot.
    }
    \label{fig:interface}
\vspace{-4mm}
\end{figure}

Although natural language feedback is cheap to collect from a chatbot's end-users, most often, feedback cannot be used directly as a training sample since feedback is usually not the answer itself, but simply contains hints to the answer.
\Cref{tab:response_samples} shows some feedback text samples.
Naive modification of feedback using heuristics like regular expressions would lead to generic responses that are ineffective in improving the dialog ability of chatbots \cite{li2016diversity}.
Additionally, writing an exhaustive set of regular expression rules is time consuming and requires extensive analysis of the data.
 Annotating data to convert feedback text to natural response is also expensive and defeats the purpose of learning from feedback text.
 
\begin{table}[t]
\centering
\footnotesize
\begin{tabular}{|l|}
\hline
you could say hey, i’m 30. how old are you?               \\ \hline
yes, i play battlefield would be a great answer \\ \hline
tell me what your favorite breakfast food is              \\ \hline
answer the question about having children!                \\ \hline
\end{tabular}
\caption{\label{tab:response_samples} Samples of feedback to the chatbot. These contain hints to the answer but they are not the answers themselves.}
\vspace{-4mm}
\end{table}
In this work, we propose a generative adversarial setup for converting such noisy feedback instances into natural, human-like responses that provide better training signals for the dialog agents.
\Cref{fig:interface} gives a bird's-eye view of our problem.
We frame this problem as a variant of text style transfer where the generator is tasked with making the feedback resemble the optimal response to the user's previous utterance and the discriminator is a classifier that distinguishes whether a given response is feedback or natural.

Our main contributions are the following:
\begin{itemize}
    \item We introduce \feedresp, a text style transfer system that converts feedback to natural responses without full supervision, thus generating additional training samples (\Cref{sec:feedresp}).
    \item We show that the training on \feedresp modified responses leads to improved accuracy of chatbots (\Cref{sec:experiments}).
    Our results also reveal that training naively on feedback doesn't help when the original chatbot is already a strong model, whereas \feedresp also helps strong models.
\end{itemize}

\section{Feedback to Natural Response  Model}
\label{sec:feedresp}
\citet{hancock2019learning} introduce a novel variant of a self-feeding chatbot in which the dialogue agent is equipped with the capability of extracting new training samples while in conversation with humans after deployment (Figure 1). The agent also employs a satisfaction module which is trained to predict how satisfied the partner is with the responses it provides. When the chatbot is engaged in a conversation where the predicted satisfaction is below a defined threshold(usually 0.5), a feedback loop is triggered where the agent requests feedback from the human user on what should have been the response. The agent then utilizes the feedback text as the target response in new training examples for the primary dialogue ranking task. \citet{hancock2019learning} show that this cost-efficient method of extracting new examples improves the chatbot's dialogue abilities. In this work, we show that naive use of the collected feedback is not necessarily a good technique and instead, we propose an approach to  better utilize the collected feedback samples. 

We pose the problem of converting feedback to resemble natural response as a text style transfer problem. We observe that feedback is more instructional and judgemental, whereas natural response is direct (answering questions) and engaging (asking questions, contains humor).
We naturalize the feedback to a response and use it as an additional training sample to improve the chatbot.

\begin{figure}[tp]
\centering     
\includegraphics[width=\columnwidth]{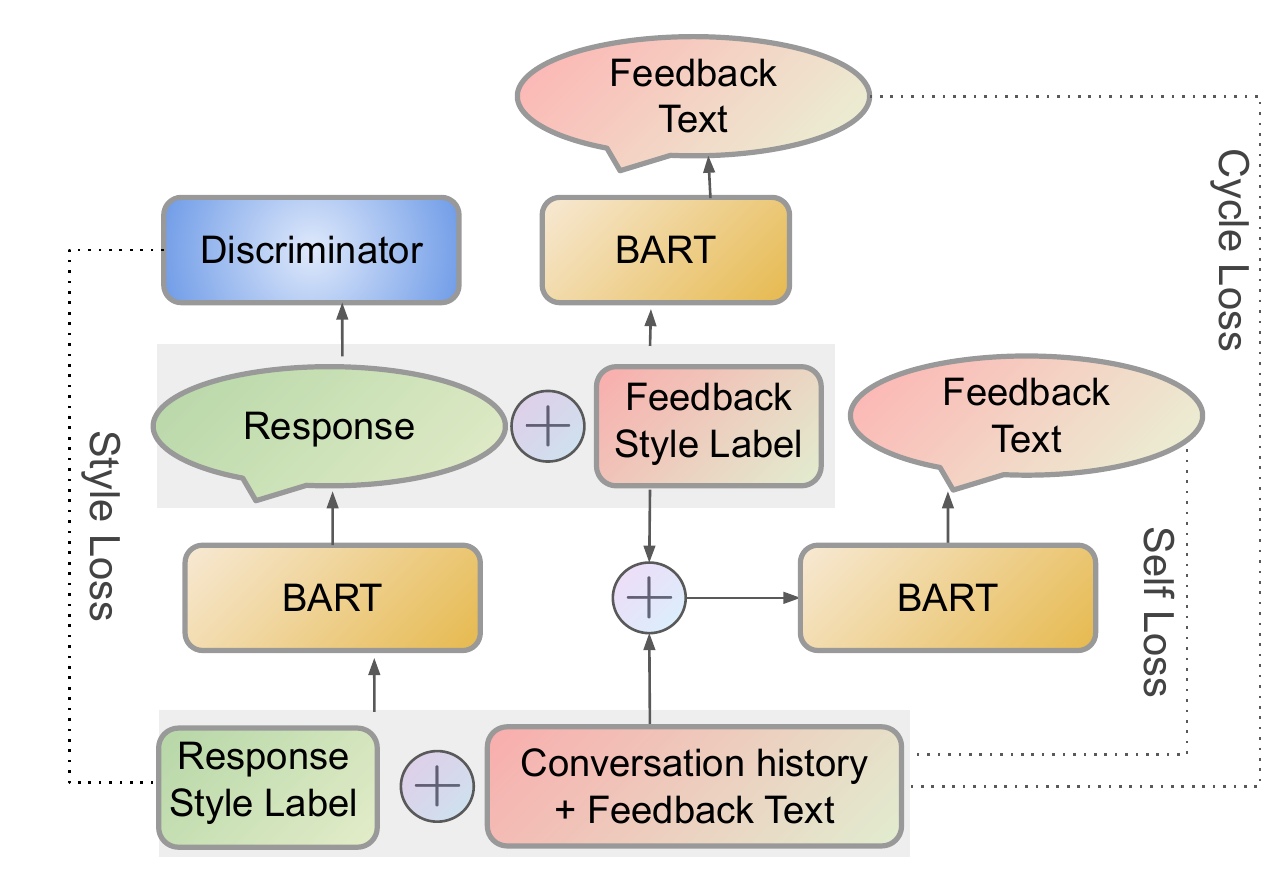}
\caption{\feedresp Architecture}
\label{fig:a}
\vspace{-4mm}
\end{figure}

A fully supervised approach to convert feedback to natural response is infeasible as we do not have paired (feedback $\leftrightarrow$ response) examples and thus we adopt an adversarial setup.
We utilize a GAN \citep{gangoodfellow} formulation where the generator modifies the feedback's style to make it seem part of a natural conversation, and in turn fool the discriminator which knows how to distinguish natural responses and feedback.
Our model, \feedresp, is shown in \Cref{fig:a}.

\subsection{Adversarial Setup}
Given an input sentence $\mathbf{x}$ (feedback or natural response) with source style $\mathbf{s}$, conversation history $\mathbf{h}$ and target style $\widehat{\mathbf{s}}$, the generator performs the mapping
\vspace{-2mm}
\begin{equation}
    g_{\theta}:(\mathbf{x}, \mathbf{h}, \widehat{\mathbf{s}})\mapsto\widehat{\mathbf{y}}
    \vspace{-2mm}
\end{equation}
Here $\widehat{\mathbf{y}}$ is the rewrite of $\mathbf{x}$ into style $\widehat{\mathbf{s}}$.
It is often the case that feedback and desired responses share many words (see \Cref{tab:pred-examples}).
 We use BART encoder-decoder initialized with pretrained weights as our generator since its denoising objective helps in copying from the input while also producing realistic sentences  \cite{lewis2019bart}.

We additionally pretrain our model under the summarization setting to extract only the response when presented with conversation history and response. 
This helps maintain brevity while still integrating details from the context in the response.
 
The discriminator is a transformer encoder network that learns to distinguish the style of feedback and natural responses.
Given an input text $\mathbf{x}$ and conversation history $\mathbf{h}$, it predicts the style class $\mathbf{c}$ of $\mathbf{x}$.
Formally, it is defined as follows:
\vspace{-4mm}
\begin{equation}
  d_{\phi}:(\mathbf{x}, \mathbf{h}) \mapsto\mathbf{c} 
  \vspace{-2mm}
\end{equation}

\subsection{\feedresp Learning}
We train \feedresp on three main objectives that help the model to reconstruct sentences when the style is not changed, change its style meaningfully and distinguish different styles.
These objectives are shown to work well in other style transfer scenarios \cite{dai2019style}.

\paragraph{Self reconstruction objective}
For the scenario where the target style is the same as the source style, we train the generator to reconstruct the sentence given as input. 
Considering the input sentence as $\mathbf{x}$, the source and the target style as $\mathbf{s}$, we minimize the negative log-likelihood loss to generate the same sentence $\mathbf{x}$ as output
\begin{equation}
    {Loss}_{self}(\theta)=-\log p_{\theta}(\widehat{\mathbf{y}}=\mathbf{x} | \mathbf{x}, \mathbf{h}, \mathbf{s})
\end{equation}
\paragraph{Cycle consistency objective}
Taking inspiration from Cycle GAN \cite{zhu2017unpaired}, we introduce a cycle consistency constraint to ensure that the model learns to preserve the meaning when it modifies the style of the original sentence.
We first transform $\mathbf{x}$ to style $\widehat{\mathbf{s}}$ to produce $\widehat{\mathbf{y}}$, i.e., $g_{\theta}(\mathbf{x}, \mathbf{h}, \widehat{\mathbf{s}})$.
 
Subsequently, we feed as input $\widehat{\mathbf{y}}$ with the target style as ${\mathbf{s}}$ and the model is trained to reconstruct the original sentence $\mathbf{x}$. 
We minimize the negative log-likelihood loss which is given by,

\begin{equation}
    {Loss}_{cycle}(\theta)=- \log p_{\theta}\left(\mathbf{y}=\mathbf{x} | g_{\theta}(\mathbf{x}, \mathbf{h}, \widehat{\mathbf{s}}), \mathbf{h}, \mathbf{s}\right)
\end{equation}
\paragraph{Style modification objective}
To ensure that the style of an input sentence $\mathbf{x}$ is changed to match the target one $\widehat{\mathbf{s}}$, we use the discriminator's confidence as training signal.
The generator wants to maximize the probability of the discriminator to classify transformed input to the target style, and therefore, we use the negative log-likelihood of the discriminator as our loss.

\begin{equation}
    {Loss}_{style}(\theta)=-p_{\phi}\left(\mathbf{c}= \widehat{\mathbf{s}} | g_{\theta}(\mathbf{x}, \mathbf{h}, \widehat{\mathbf{s}}\right))
\end{equation}

\subsection{End-to-end training}

 The discrete nature of sampling and non-differentiability of the argmax operator prevents gradient backpropogation. 

Following \citet{dai2019style}, we consider the softmax distribution produced by the generator, $g_{\theta}$ as the `soft' generated sentence and use it as input for further downstream networks to maintain differentiability. 
\section{Experimental Setup}
\label{sec:experimental-details}

In \feedresp, the optimizer for both the generator and discriminator is AdamW. The learning rate of generator is 5e-6 while the learning rate of discriminator is 1e-4. The discriminator uses 4 stacked transformer layers and 4 attention heads. The token embedding size, style embedding size, positional embedding size and hidden size are all 256. For the BART \cite{lewis2019bart} generator, we use the implementation from HuggingFace \cite{Wolf2019HuggingFacesTS} and initialize the model with pretrained weights from the CNN/Daily Mail summarization task. Due to the characteristics of human response(refer Appendix \ref{dataset-statistics}), we limit the length of text generation to a maximum of 50 words and impose a repetition penalty of 2.0 to improve diversity of output.\bigbreak
While evaluating the effectiveness of the modified feedback responses, we use two implementations of dialog agents provided by ParlAI \cite{miller2017parlai},  \biencoder and \polyencoder. \biencoder has two transformer layers and 2 attention heads. The optimizer is Adamax with learning rate of 0.0025. \polyencoder uses 12 transformer layers and 12 attentions heads. The optimizer is Adamax with learning rate of 5e-05.\bigbreak
The hyperparmeters for the best performing model are arrived at by random sampling and subsequently verifying the outputs using human evaluation to rate the outputs from the style transfer task. The entire list of hyper-parameters is listed in the Table~\ref{tab:hyperparameters}.
\section{Experiments}
\label{sec:experiments}

Our goal is to test whether feedback helps improve the chatbot.
To do this, we compare models trained on conversational data with and without feedback data.
Below we describe the chatbot evaluation setting, our datasets, the main models and different settings of these models with and without feedback.

\subsection{Chatbot evaluation task and metrics}
Following \citet{hancock2019learning}, we choose PersonaChat \cite{zhang2018personalizing} as the main evaluation dataset.
This dataset consists of human-human conversations collected using crowdsourcing where each crowdworker takes a persona.
Since persona representation is a challenging research problem on its own, \citeauthor{hancock2019learning} ignore the persona and just use the conversations to train chatbots and we follow the same approach.
At test time, the model is presented the conversation history and 20 candidate responses and the model has to pick the correct response. Thus, we use HITS@1/20 metric for evaluation.

\subsection{Feedback data}
We use the feedback data collected by \citet{hancock2019learning} as this removes orthogonal factors such as differences in chatbot interfaces and annotation framework etc. which are not the focus of this work.
\citeauthor{hancock2019learning} collected this feedback by deploying bi-encoder chatbots (Section ~\ref{sec:models}) trained on varying levels of training data and making it converse with crowdworkers.
Whenever the bot's response is not satisfactory, natural language feedback is collected from the crowdworker.

The data thus collected contains 60k human-bot turns, of which the last turn is always the feedback.

\subsection{Chatbot Models} 
\label{sec:models}
Given the conversation history and several candidate responses, the chatbot is trained to rank the correct candidate on the top.
We use the following models as our chatbots.
\vspace{0.2em}

\noindent \textbf{\biencoder} \cite{hancock2019learning,humeau2020poly} contains two transformers, one for summarizing the conversation history and the other to summarize candidate responses to embeddings. 
The response with highest similarity is taken as the best candidate response.
\vspace{0.2em}

\noindent \textbf{\polyencoder} \cite{humeau2020poly} summarizes a context and candidate responses into several embeddings.
In order to contextualize context and candidates together, it performs a cross-encoder attention on the summary embeddings and scores each candidate.

\subsection{Feedback-based Models}
\label{sec:feedback_models}
We train and test the above models in the following settings.
\vspace{0.2em}

\noindent \textbf{\nofeed}:  The model is trained only on human conversations.
\vspace{0.2em}

\noindent \textbf{\feedback}: We train on the combination of human conversations and unmodified feedback data.
This setting is similar to \citet{hancock2019learning}.
\vspace{0.2em}

\noindent \textbf{\heuristic}: We design and use six regular expression rules based on the frequent patterns in the data that convert feedback to plausible dialog responses (see \Cref{app:regex-rules})
and train the chatbot models on human conversations along with the modified feedback.
\vspace{0.2em}

\noindent \textbf{\feedresp}: We use our main model (\Cref{sec:feedresp}) to modify feedback to natural responses and train the chatbot models on modified feedback along with human conversations.

\begin{table}[tp]
\footnotesize
    \begin{center}
    \begin{tabular}{lcc}
    \toprule 
    {\bf Model}          &   {\bf Development} &  {\bf Test}\\
   \midrule
   \multicolumn{3}{c}{\biencoder chatbot} \\
   \midrule
  \bf   \nofeed                & 49.03 (0.66)   & 49.49 (0.49) \\
  \bf   \feedback           & 49.27 (1.06)    & 49.97 (1.30) \\
  \bf   \heuristic      & 48.85 (0.70) & 49.85 (0.72) \\
  \bf  \feedresp       & 50.84 (0.50)     & 51.32 (0.43) \\
   \midrule
   \multicolumn{3}{c}{\polyencoder chatbot} \\
   \midrule
  \bf   \nofeed        & 73.35 (0.70)  &   69.94 (0.37)	 \\
  \bf   \feedback      & 72.63 (0.14)  &    68.48 (0.64) \\
   \bf  \heuristic     & 72.65(0.35)   &    68.83(0.31) \\
  \bf  \feedresp       & \bf 78.14 (0.40)  &   \bf 75.96 (0.80) \\
    \bottomrule
    \end{tabular}
    \end{center}
    \caption{Hits@1/20 of models on \personachat. 
    Naive and heuristic use of feedback results in marginal improvement or hurts performance, whereas \feedresp modified feedback gives large improvements. 
    The variances across three different runs are also shown.}
    \label{tab:results} 
    \vspace{-2mm}
\end{table}

\section{Results and Discussion}
The experimental details of the model variants are described in \Cref{sec:experimental-details}.
\Cref{tab:results} shows the average HITS@1/20 of all models on the \personachat validation and test sets over 3 runs.
We were able to replicate results of \citet{hancock2019learning} which show that \biencoder performance improves slightly (+0.48 on test) when \feedback is used.
\heuristic edits to feedback don't help while
\feedresp responses improve the results higher than \feedback and also have less variance.
Coming to \polyencoder, it is a much stronger chatbot than \biencoder.
We see that naive use of \feedback or \heuristic deteriorates the performance of \polyencoder while \feedresp emerges a clear winner with +6.0 point improvement on the test set over \nofeed.

\begin{table}[tp]
\footnotesize
\begin{tabular}{p{4.5cm}ll}
\toprule
Example & Freq. & Acc. \\
\midrule
\multicolumn{3}{c}{Modification type: Rewrite} \\
\midrule
F: tell me about your favorite show & 18.5\% &  81\% \\
F2R: I love watching TV shows and sitcoms like friends \\
\midrule
\multicolumn{3}{c}{Modification type: Remove} \\
\midrule
 F: you could’ve said, yes the sugar cinnamon kind is my favorite  & 40\% & 68.7\%  \\
 F2R:  yes the sugar cinnamon kind is my favorite \\
\midrule
\multicolumn{3}{c}{Modification type: Retain} \\
\midrule
F:  the temperature is hot & 41.5\% & 74.6\% \\
F2R: the weather is hot  \\
\bottomrule
\end{tabular}
\caption{Statistics of different modification types based on 200 random feedback texts.
F stands for feedback, and F2R is the response of \feedresp model.
Freq. indicates the frequency of the modification type, and Acc. the accuracy of \feedresp on each type. \Cref{app:feed2resp examples} lists additional examples of modified feedback responses.}
\label{tab:ling-phenomena}
\vspace{-2mm}
\end{table}
\begin{figure*}
    \centering
    \includegraphics[width=\textwidth]{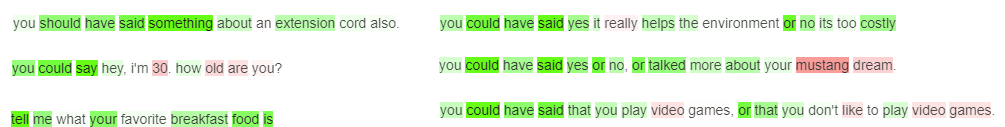}
    \caption{Attention maps for Feedback responses. Words such such \textit{you should have}, \textit{you could have}, \textit{tell me} heavily influence the discriminator to classify it as feedback and hence the generator learns to remove such words to fool the discriminator. Darker shades of green mean higher attention scores and shades of red mean lower attention scores.}
    \label{fig:my_label}
\end{figure*}
\paragraph{Feed2Resp analysis}
We randomly sample 200 feedback responses from \feedresp to determine the kind of modifications the model performs (\Cref{tab:ling-phenomena}).
We observe three main types of modifications --- \textit{Rewrite, Retain and Remove}.
\rewrite is when the feedback implies an hint to the answer but not the answer itself.
\remove is when the feedback contains the answer with extraneous words that have to be removed.
\retain are cases where the model copies or paraphrases the feedback.
Among these, \remove has the lowest accuracy of modification.
Upon inspection, we find that these are the cases which require multiple removals.
For example, for \textit{You should reply with either yes or no}, the model predicts \textit{yes or no} together instead of either one of them. 
Additionally, we visualize the attention maps of the discriminator to observe which words contribute most to the classification decision of the discriminator (\Cref{fig:my_label}).  The discriminator learns to distinguish feedback from normal dialog responses due to the presence of sequences like \textit{you could have}, \textit{you should have}, \textit{tell me}, etc. 
Thus the generator learns to remove such extraneous sequences and make the feedback seem like plausible responses.
We present a sample of modified outputs of \feedresp in \Cref{app:feed2resp examples}.

\section{Conclusion}
In this work, we show that while chatbots can be improved using natural language feedback, converting feedback to natural responses that fit in the conversation outperform the naive usage of feedback. 
We presented \feedresp, a generative adversarial model, that converts feedback to natural responses without requiring manually annotated parallel data.
Our results show that \feedresp results in a 6~point improvement for the \polyencoder chatbot, an already powerful dialog ranking agent. 
This is a strong result as HITS@1/20 is a tough metric to improve upon \citep{hancock2019learning}.

Our work joins the class of models that use natural language feedback to improve different tasks, e.g., image captioning \cite{NIPS2017_7092}, classification \cite{srivastava2017joint,hancock-2018-training,murty2020expbert}.
While these methods use feedback for reward shaping or feature extraction, we use feedback to produce correct response using adversarial learning.
We pose this problem as a style transfer problem inspired from the style transfer literature \citep{shen2017style,xu2018unpaired,li2018delete,lample2019,dai2019style}. 
While these focus on studying the stylistic attributes of sentences, e.g, sentiment, we explore this problem in the context of improving chatbots.

\section{Acknowledgements}
We thank Yue Dong for her multiple helpful discussions during the course of this project. We also thank Sandeep Subramanian for his insightful guidance at a crucial stage of this work. This research was enabled in part by computations support provided by Compute Canada (www.computecanada.ca).
The last author is supported by the NSERC Discovery Grant on \textit{Robust conversational models for accessing the world's knowledge}.

\bibliographystyle{acl_natbib}
\bibliography{emnlp2020}

\appendix

\clearpage

\section{Dataset Statistics}
\label{dataset-statistics}
\par We are going to validate our approach on the chatbot's performance using \personachat \cite{zhang2018personalizing} dialogue dataset  and Human-Bot feedback dataset \cite{hancock2019learning}. 
Table \ref{tab:datasets} reports the size of each dataset, all of which are available via ParlAI.\footnote{https://parl.ai/projects/self\_feeding/}

\begin{table}[ht]
    \begin{center}
    \setlength\tabcolsep{1.5pt}
    \begin{tabular}{lrrrr}
    \hline
    Task           & Train    & Valid  & Test & Total \\
    \hline
    \dialogue {\tiny{\textsc{(Human-Human)}}}   & 131438   & 7801   & 6634 & 145873 \\
    \feedback {\tiny{\textsc{(Human-Bot)}}}     & 60000    & 1000   & 1000 & 62000 \\
    \hline
    \end{tabular}
    \end{center}
    \caption{\label{tab:datasets} The number of examples used in our experiments by task and split. Note that the HH \dialogue examples come from the \personachat \ dataset, HB \feedback \ examples from \citet{hancock2019learning}}
\end{table}

To train the \feedresp model, we take the entire \feedback dataset and an equal number of randomly chosen samples from the \dialogue dataset. We them use a train-dev-test split of 0.8:0.1:0.1 for training and evaluation of the model.

\begin{table}[!htbp]
    \begin{center}
    \setlength\tabcolsep{1.5pt}
    \begin{tabular}{lrrrr}
    \hline
    Task           & Train    & Valid  & Test & Total \\
    \hline
   Style Transfer &  96000 & 12000 & 12000&120000 \\
    \hline
    \end{tabular}
    \end{center}
    \caption{\label{tab:datasets} The number of samples per split used in our style transfer experiment. We take an equal number of samples from the \feedback and \dialogue datasets and randomly shuffle them to create train, validation and test splits. The number of samples of each class in all the splits are ensured to be evenly distributed.}
    \end{table}

\begin{table}[!htbp]
    \begin{center}
    \setlength\tabcolsep{1.5pt}
    \begin{tabular}{lrrr}
    \hline
    Statistic           & Human-Human    & Feedback \\
    \hline
    \#Words in context {\tiny{\textsc{(mean)}}}     & 79   & 13 \\
    \#Words in context {\tiny{\textsc{(median)}}}   & 77    & 6 \\
    \#Words per turn {\tiny{\textsc{(median)}}}     & 10.7    & 7.1 \\
    \#Words per turn {\tiny{\textsc{(mean)}}}       & 11   & 6 \\
    \hline
    \#Turns {\tiny{\textsc{(mean)}}}                & 4   & 1.5 \\
    \hline
    \end{tabular}
    \end{center}
    \caption{\label{tab:datasets} The number of words  per context and per turn. The second part is the average number of turns in a conversation.}
\end{table}
We examine the average number of turns and words in dialogues from the the feedback and human-human conversation distributions. We see that on an average, the dialogues in the feedback distribution have fewer number of turns than in human-human conversations. The average number of words per turn is also fewer on average.
\section{Preparation of Training Data}

\par We use the dataset provided by \citet{hancock2019learning}, which is a cleaner version of \personachat dataset and comes with a new crowdsourced test set. We sample an equal number of examples from the \dialogue dataset, giving them a label  0, and \feeback dataset, giving them a label of 1. The final response are combined with last n turns with an delimiter [RES]. Typically, n=2 turns are used for each conversation example. Conversation turns are separated with delimiter tokens [P1] or {[P2]}.
\section{\feedresp examples}
\label{app:feed2resp examples}
Here we include several examples of predictions from different models in Table \ref{tab:pred-examples} .

\section{Computing Infrastructure and Runtime of Experiments}
All experiments are conducted on Nvidia V100 GPUs. Average runtime of experiments is listed in Table \ref{tab:runtime} and assume running experiments on a single Nvidia V100 GPU.

\begin{table}[!ht]
    \begin{center}
    \setlength\tabcolsep{1.5pt}
    \begin{tabular}{lr}
    \hline
    Task           & Approximate Training Time \\
    \hline
   \biencoder  &  16H \\
   \polyencoder & 16H \\
   \feedresp & 48H \\
    \hline
    \end{tabular}
    \end{center}
    \caption{\label{tab:runtime} Average runtime(in hours) of various experiments conducted}
\end{table}

\section{Regular Expressions to modify \feeback}
\label{app:regex-rules}
As described in Section~\ref{sec:feedback_models}, we use the combination of following regular expressions to strip the filler words and extra choices in feedback data: 

\begin{verbatim}
r"you could have|you should have|
you could|you should"
r"^said|^saying|^say|^tell |^told 
|^admit |asked |^ask |^answer 
|^answered |^talked |^talk "
r"^about|^me|^that"

r"if|whether|not" "you are " to "i am "
"your "to "my "
"you\'ve " to "i\'ve "
"you were" to "i was"
"you " to "i "
    
\end{verbatim}

\begin{table*}[ht]
    \begin{center}
    \renewcommand{\arraystretch}{0.9}
    \setlength\tabcolsep{10pt}
    \begin{tabular}{lllllll}
    \toprule
    {\bf Model } & {\bf layers} & {\bf heads} & {\bf optimizer} & \bf{learning rate} & {\bf \#parameters} \\
    \midrule
    BART Large        & 12   & 16 & AdamW &  5e-6   & 406M\\ %
    Discriminator    & 4   & 4 & AdamW &  1e-4  & 16M\\ %
    \midrule
    \biencoder                 & 2   & 2  & Adamax &  2.5e-3 & 23M\\ %
    \midrule
    \polyencoder               & 12   & 12 & Adamax &  5e-05  & 256M\\ %
    \bottomrule
    \end{tabular}
    \end{center}
    \caption{The hyperparameters used to obtain the results in Table~\ref{tab:results}.}
    \label{tab:hyperparameters}
\end{table*}

\begin{table*}[ht]
\footnotesize
\begin{center}
\begin{tabular}{lp{\columnwidth}}
\toprule
Conv. Hist.             & [H] It makes more sense to order food online. There are better deals.  [B] good choice. what is your favorite food?  [H] pizza. what is yours?\\
Feedback & you could have spoken about your favorite food \\
\feedresp       & my favorite food is pasta.                              \\ 
\midrule
Conv. Hist.                                & [H] hello [B] hi. how are you doing? [H] what do you do for a living? \\ 
Feedback & tell me a job or career \\
\feedresp                       &   i am a teacher. \\ 
\midrule
Conv. Hist.                                & [H] hello [B] hello [H] do you like to watch movies? \\ 
Feedback & yes or no \\
\feedresp                       &   yes i like movies or no i hate movies. \\ 
\midrule
Conv. Hist.                                & [H] I tried to climb a tree and broke my ankle. \\ 
Feedback & that sounds painful \\
\feedresp                       &   i am sorry. that sounds painful. \\ 
\midrule
Conv. Hist.                                & [H] hello [B] hi. [H] do you exercise? \\ 
Feedback & yes, i do exercise. \\
\feedresp                       &   yes, i do exercise and go jogging. \\ 
\midrule
Conv. Hist.                                & [H]  what do you think about wintertime? \\ 
Feedback & you could say that you like snow or you think it is too cold. \\
\feedresp                       &  you like snow or you think it is too cold.  \\ 
\midrule
Conv. Hist.                                & [H] hello [B] hello [H] have you ever ate chinese food? \\ 
Feedback & yes i like chinese. \\
\feedresp                       &   yes i like chinese noodles and dumplings. \\ 
\midrule
Conv. Hist.                                & [H] who is the greek goddess of love?  [B] yes , i lived in south florida for years [H]] sorry, i meant who is the greek goddess of love? \\ 
Feedback & venus \\
\feedresp                       &   venus is a very good player. he has a lot of experience playing with the likes of jurgen klinsmann. \\ 
\bottomrule
\end{tabular}
\end{center}
\caption{Examples of transformed feedback by our model, Feed2Resp, resulting in specific responses.}
\label{tab:pred-examples}
\end{table*}

\end{document}